\documentclass[10pt,twocolumn,letterpaper]{article}

\usepackage{cvpr}              

\usepackage{graphicx}
\usepackage{amsmath}
\usepackage{amssymb}
\usepackage{booktabs}

\usepackage[pagebackref,breaklinks,colorlinks]{hyperref}

\usepackage[capitalize]{cleveref}
\crefname{section}{Sec.}{Secs.}
\Crefname{section}{Section}{Sections}
\Crefname{table}{Table}{Tables}
\crefname{table}{Tab.}{Tabs.}

\def\cvprPaperID{*****} 
\def\confName{CVPR}
\def\confYear{2022}

\begin{document}

\title{\textbf{MegaPortrait}: Revisiting Diffusion Control for High-fidelity Portrait Generation}

\author{Han Yang\textsuperscript{1,2}\,\,\,\,\,\,\,\,\,\,  Sotiris Anagnostidis\textsuperscript{1}\,\,\,\,\,\,\,\,\,\,  Enis Simsar\textsuperscript{1}\,\,\,\,\,\,\,\,\,\,  Thomas Hofmann\textsuperscript{1} \\
\centerline{
\textsuperscript{1}ETH Zurich\,\,\,\,\,
\textsuperscript{2}ZMO AI Inc.}\\
{\tt\small hanyang@ethz.ch}
}

\begin{figure}[htb]

\twocolumn[{
\renewcommand\twocolumn[1][]{#1}%
\maketitle
\vspace{-32pt}

\begin{center}
  \centering
  \includegraphics[width=1\textwidth]{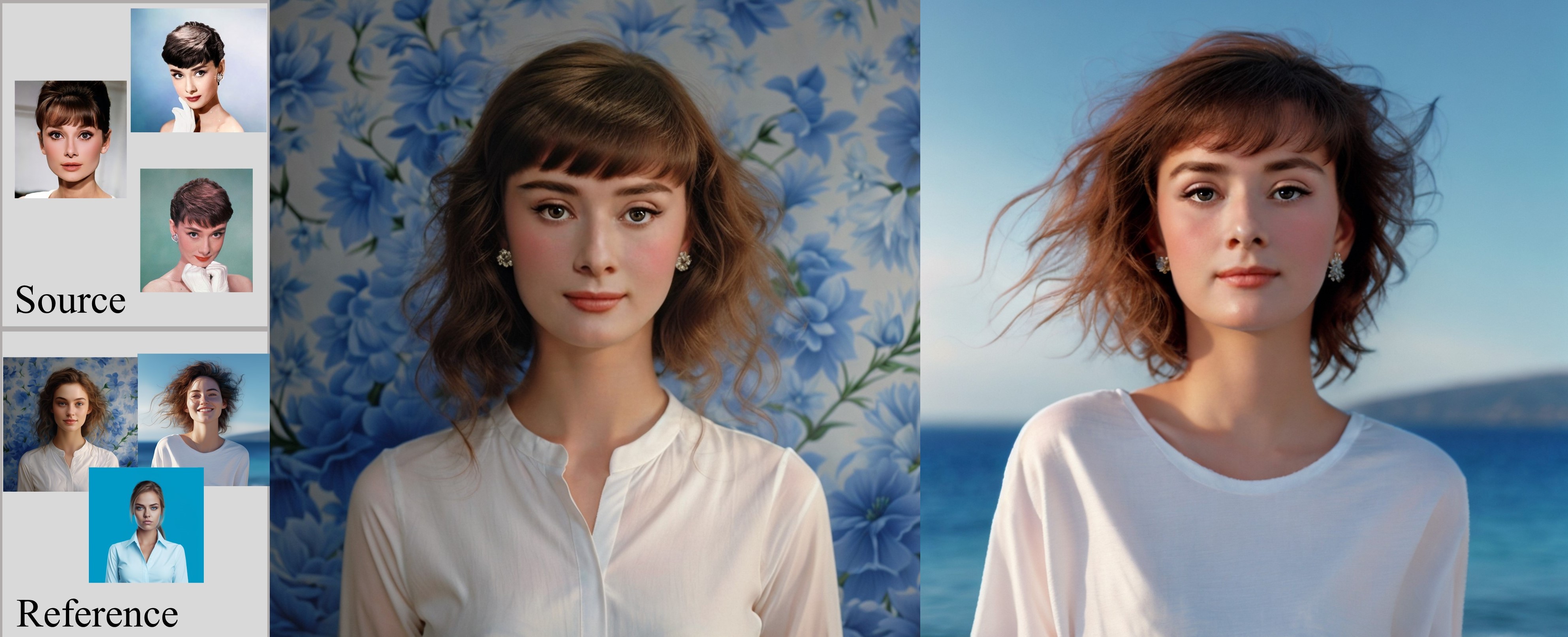}
  \vspace{-20pt}

\end{center}

\caption{Our novel approach, \textbf{MegaPortrait}, produces results of exceptional quality. Leveraging the person ID provided by the source image, our method adopts the reference image as style and pose reference. MegaPortrait generates high-fidelity results and seamlessly integrates the source individual's features with specified styles and poses extracted from reference image.}
\label{fig:teaser}
}]
\end{figure}

\begin{abstract}

We propose MegaPortrait, an innovative system tackling the challenge of personalized portrait image creation in computer vision. Through three meticulously designed modules – Identity Net, Shading Net, and Harmonization Net – we achieve a delicate balance between identity preservation and artistic adaptation. Leveraging a customized model fine-tuned with source images, Identity Net generates learned identity, while Shading Net re-renders portrait images using extracted representations. Harmonization Net seamlessly fuses pasted faces and the reference image's body, ensuring coherent final results. Our approach, utilizing off-the-shelf Controlnets, surpasses state-of-the-art AI portrait products in both identity preservation and image fidelity, establishing MegaPortrait as a significant advancement in the field. Without additional so-called novelties, MegaPortrait introduces a simple but effective system design in tackling a well-established problem and we compare our results with the state-of-the-art AI portrait methods as well as products to demonstrate the superiority of our proposed system in identity preserving and image fidelity.

\end{abstract}

\begin{figure*}[th]
  \centering
   \includegraphics[width=1\linewidth]{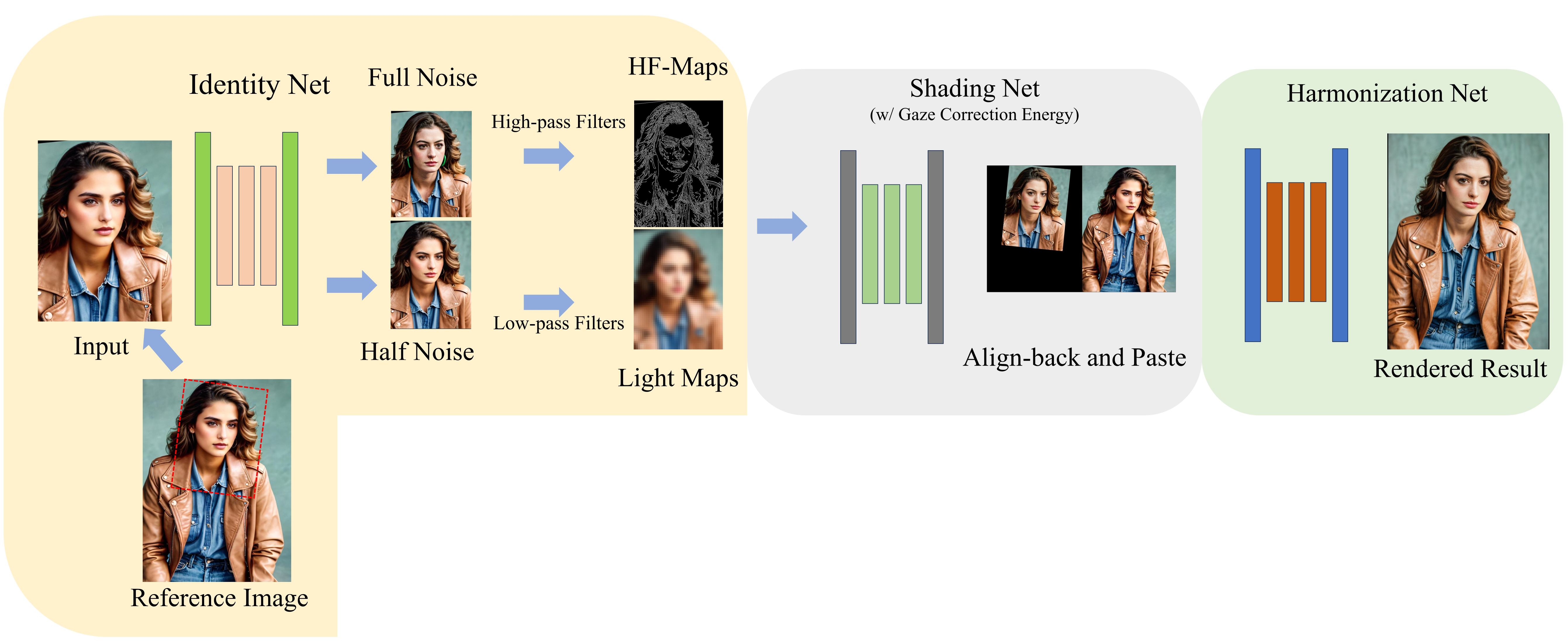}

   \caption{The overall pipeline of our MegaPortrait which consists of three modules: Identity Net, Shading Net and Harmonization Net. The Identity Net generates an image pair with a different identity-shading trade-off, and we extract the light maps with low-pass filters from the more-reference-like image and the HF-Maps from the more-source-like image. Then the Shading Net takes these two control conditions to synthesize the geometrically correct stylized image and paste back to the reference image for final harmonization, conducted by the Harmonization Net.}
   \label{fig:pipeline}
\end{figure*}

\section{Introduction}
The development of AI-driven portrait generation within the framework of diffusion models encounters multifaceted challenges necessitating innovative solutions. The central issue revolves around striking a delicate balance: preserving subject identity while enabling effective generalization across diverse artistic styles. This persistent challenge has historically impeded progress in the field. For generative portrait models to succeed, they must seamlessly adapt to various cues and preferences, ensuring that the resulting images faithfully capture the unique characteristics of each subject while embodying the intended style or mood. Addressing this dilemma directly, our proposed method, ``MegaPortrait", offers a solution by decoupling shading or color information from the geometry through our split-and-merge pipeline. This approach allows us to generate coherent results aligned with the style reference image, overcoming the longstanding obstacle of balancing subject identity and artistic expression.

To shed insight on this challenging problem, identity and style trade-off,  instead of training a diffusion network from scratch to directly model the mapping relationship, we try to build a portrait generation system by adopting different controlnets~\cite{controlnet} and carefully picking the input conditions. Our system can be recognized as a free-lunch application of controlnet and provides new insights on designing next-level generative applications. Current AI portrait designs basically focus on training better identity encoder such as ~\cite{FastComposer,IP-Adapter,InstantID}, but fail to maximize the conventional generative power in pretrained diffusion models~\cite{LatentDiffusion}. The capabilities in such design are limited by the design choices of encoders.

Inspired my portrait relighting~\cite{totalrelighting}, we try to decouple the shading and structure first and fuse them together. Specifically, the comprehensive MegaPortrait pipeline consists of three carefully designed modules: Identity Net, Shading Net, and Harmonization Net, which are collectively designed to produce high-fidelity and visually compelling personalized portrait images. In the initial stage, the identity network dynamically generates a pair of images, each embodying a different identity coloring trade-off. From this pair of images, we extract the illumination map from the image through a low-pass filter wise, exhibiting a more reference-like quality, while deriving a high-frequency map (HF-Map) from the image with a more source-like appearance. This dual extraction process ensures a meticulous rendering of subject identity and shadow characteristics.

The subsequent Shading Net module uses these extracted light maps and HF-Map as control conditions to perform the synthesis process and generate geometrically correct and stylized portrait images. The images were carefully crafted to balance shadow and identity, then seamlessly pasted back onto the reference image. The final harmonization step is expertly handled by Harmonization Net, which integrates the composite portrait with the reference image, ensuring the final result is cohesive and visually harmonious. This complex orchestration of modules in the MegaPortrait pipeline embodies our commitment to achieving a delicate balance between preserving personal identity and adapting to different artistic styles.

To summarize our contributions: 1) We propose a principled pipeline, MegaPortrait, which effectively addresses the identity-style dilemma in AI-driven portrait generation. By adopting different controlnets and carefully selecting input conditions, our system represents a free-lunch application of controlnet, providing new insights for designing next-level generative applications. 2) To shed light on the challenging identity and style trade-off problem, we introduce a novel approach of decoupling shading or color information from geometry through our split-and-merge pipeline. This technique, inspired by human relighting practices, is the first of its kind applied in realistic AI portrait generation, offering a unique perspective on overcoming the intricacies of the mapping relationship. 3) Our comprehensive MegaPortrait pipeline, consisting of Identity Net, Shading Net, and Harmonization Net, produces high-fidelity and visually compelling personalized portrait images. By surpassing research methods and top products such as Remini, both quantitatively and qualitatively, our results demonstrate the effectiveness of our approach in maximizing generative power within pretrained diffusion models.

\begin{figure*}[th]
  \centering
   \includegraphics[width=1\linewidth]{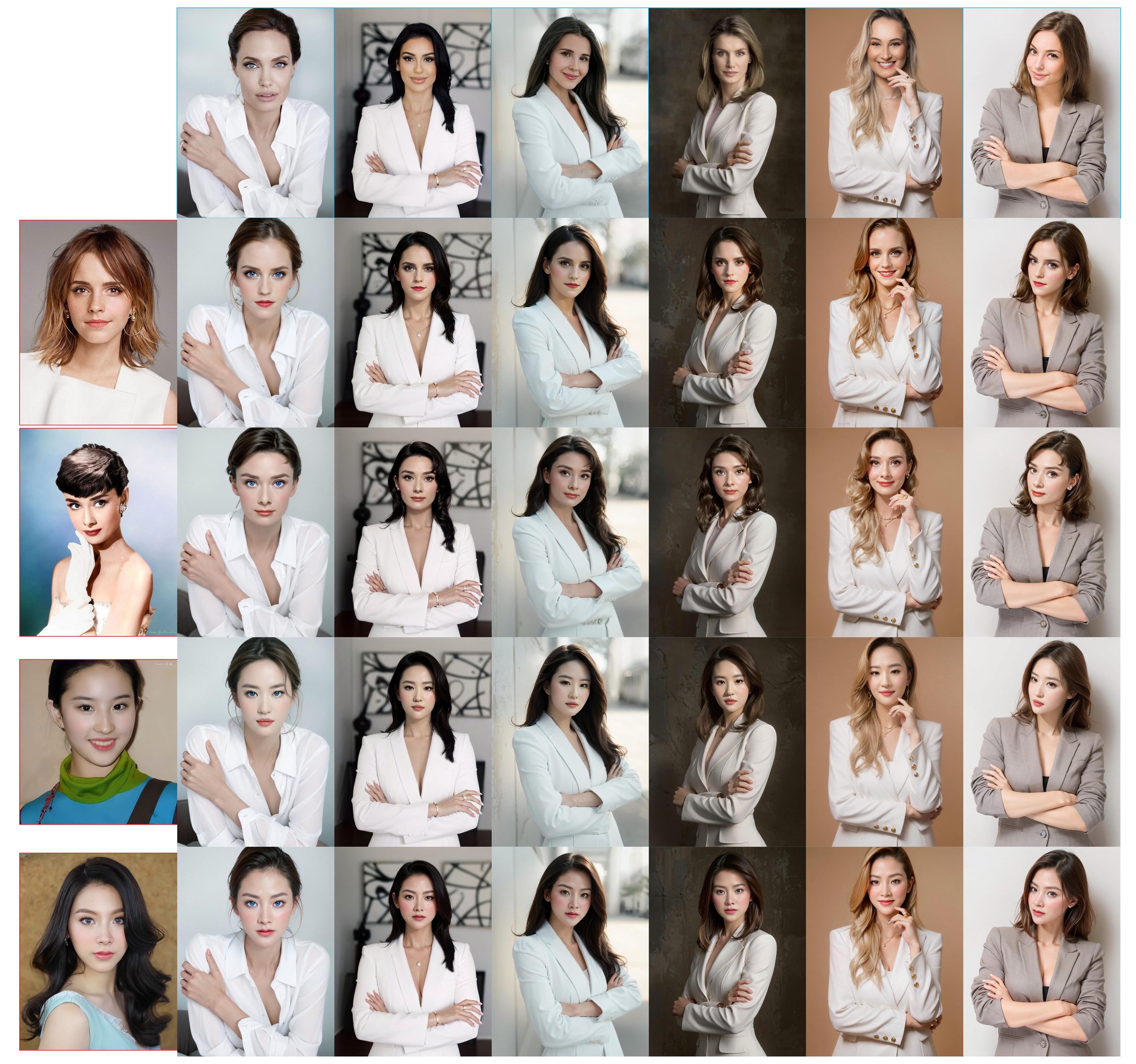}

   \caption{The cross extensive results of MegaPortrait with various source IDs and reference styles. The reference images are blue-boxed and the source images are red-boxed.}
   \label{fig:multiple}
\end{figure*}

\section{Related Works}
\subsection{Text-to-Image Generation} From generating only birds with low quality such as AttnGAN~\cite{AttnGAN} to the capability of creating photo-realistic and content-rich results, text-to-image generation has become the most remarkable advancement with the law of scalability. Stable diffusion~\cite{LatentDiffusion} or latent diffusion is the key trigger to stimulate the progress of this area, with the industry-level dataset Laion5B~\cite{Laion5B}. The benefits of diffusion models~\cite{DDPM,LatentDiffusion} is obvious compared to Generative Adversarial Networks (GANs)~\cite{AttnGAN}, which suffers instability and mode collapse; GANs inherently struggles to fit tremendous data and requires tiresome parameter tuning and discriminator design. 
\subsection{Personalized Image Generation}
Personalized image generation which targets to generate images with consistent customized appearance has achieved remarkable progress. In GANs-era, GAN-Inversion-based techniques such as PTI~\cite{PTI} can only edit well-aligned images in the GAN latent space. With the powerful ability of diffusion models, dreambooth~\cite{Dreambooth} achieves photo-realisitic results by fine-tuning the text-to-image models with user-specified images. LoRa~\cite{Lora} uses low-rank matrices to shrink the trainable parameters, which can be recognized as strongly-regularized dreambooth method. Dreambooth-like methods are usually confronted with over-fitting problems which generalize poorly on various prompts. 
\subsection{Human-centric Image Generation}
A classic methods is Faceswap~\cite{insightface} which only swaps the facial features without really generating the source identity. Pioneering diffusion methods revolutionize the basic logic by generating exactly the source person according the reference style or prompt.
 With the help of diffusion models, achieving photo-shooting-level human portrait generation is now possible. Dreambooth\cite{Dreambooth} and LoRa\cite{Lora} are capable of generating identity-preserving results, but struggle to satisfy the aesthetic demand of various photo-shooting styles. Encoder-based methods such as IP-Adapter~\cite{IP-Adapter} and Fastcomposer~\cite{FastComposer} improve the aforementioned problems by using image encoder to extract the identity of the source images to ensure generatlization abilitity, but they fail to follow specific style reference. Directly feeding style reference as image prompt causes id-mixing problem according to \cite{InstantID}. Remini~\cite{remini} is the bestseller AI Portrait product in app-store and google play, please try to download this app to see product-ready results.
\subsection{EBMs and Diffusion}
Diffusion-models~\cite{DDPM} generate images by denoising the noised images, and the noise predicted can be also recognized as score~\cite{score-matching} or gradient of energy of Energy-based Models (EBMs)~\cite{EBM}. Under the perspective of score or energy, a classic framework of classfier-free guidance~\cite{classifier-free} is proposed to generate high-fidelity images without explicit classifier guidance by compositing the diffusion score into conditional distribution by Bayesian theorem. Safe Latent Diffusion~\cite{SafeLatent} further extends this framework to eliminate the harmful information from the synthesis process, which achieves better alignment than negative prompting.

\section{Method}
We give a pipeline of our proposed method in Fig.~\ref{fig:pipeline}. By revisiting the diffusion control condition scheme~\cite{controlnet}, we successfully achieve state-of-the-art AI portrait generation quality compared to the most famous AI portrait applications and products. Motivated by the common practices in portrait relighting~\cite{totalrelighting}, we try to extract the color clues from the reference image by using similar low-pass filtered light-maps techniques.
\subsection{Geometry and Shading}
The previous dilemma of shading (color) and geometry (structure) conflict raises great challenge of decoupling the intrinsics (as ``intrinsic decomposition"~\cite{intrinsic_decomposition}) of a portrait. Credited to the power of diffusion model, we managed to ameliorate this problem by involving two representations: HF-Maps and Light Maps. Convolved Light Maps~\cite{totalrelighting} are commonly used as replacement of real physically-based rendering light map, which area generally derived from image down-sampling, while HF-Maps are the high-frequency details extracted by High-pass Filters.  

\noindent\textbf{HF-Maps}\quad To generate the HF-Maps, the identity Net generates a Full noise portrait. Here, we denote ``full noise" by using random noise as initialization to generate the ID-reference. Given the id-reference, we extract the HF-Maps by applying high-pass filters to get the geometry reference.

\noindent\textbf{Light Maps}\quad Motivated by portrait relighting~\cite{totalrelighting}, we extract the convolved light maps by applying image downsampling using gaussian blur and pixel sampling. We introduce the diffuse clues into the condition to give the shading conditions. Practically, we add noise to the encoded input image to the timestamp 400 and 700 out of 1000 to find a trade-off of identity and shading clues and extract the light maps. The design Low-pass filters concerns identity corruption (preventing ID pollution from reference ID) and shading preservation.

\subsection{Gaze Correction Guidance}
Due to the over-fitting nature of the customized model, the gaze of the generated portrait is lifeless and dull compared to real portrait. Since the prompt alignment degenerates along with the over-fitting process, we cannot control the gaze as freely as the original model.  Although hard to notice quantitatively, the gaze problem is severe as in Fig.~\ref{fig:gaze}; without gaze guidance, the synthesized person cannot look at the direction as the prompt guided (``looking at the camera" here).

The diffusion score $\epsilon_{\theta}(z_{\lambda}, c) \approx -\sigma_{\lambda} \nabla_{z_{\lambda}} \log p(z_{\lambda}|c)
$ is modified with an auxiliary classifier in the classifier-guidance setting:
\begin{equation}
    \tilde{\epsilon}_{\theta}(z_{\lambda}, c) \approx - \sigma_\lambda\nabla z_{\lambda} [\log p(c|z_\lambda) + w \log p_\theta(c|z_{\lambda})],
\end{equation}
where $w$ is the weight to control the classifier guidance. The modified score $\tilde{\epsilon}_{\theta}(z_{\lambda}, c)$ replaces the original score in the diffusion process.

\begin{figure*}[th]
  \centering
   \includegraphics[width=1\linewidth]{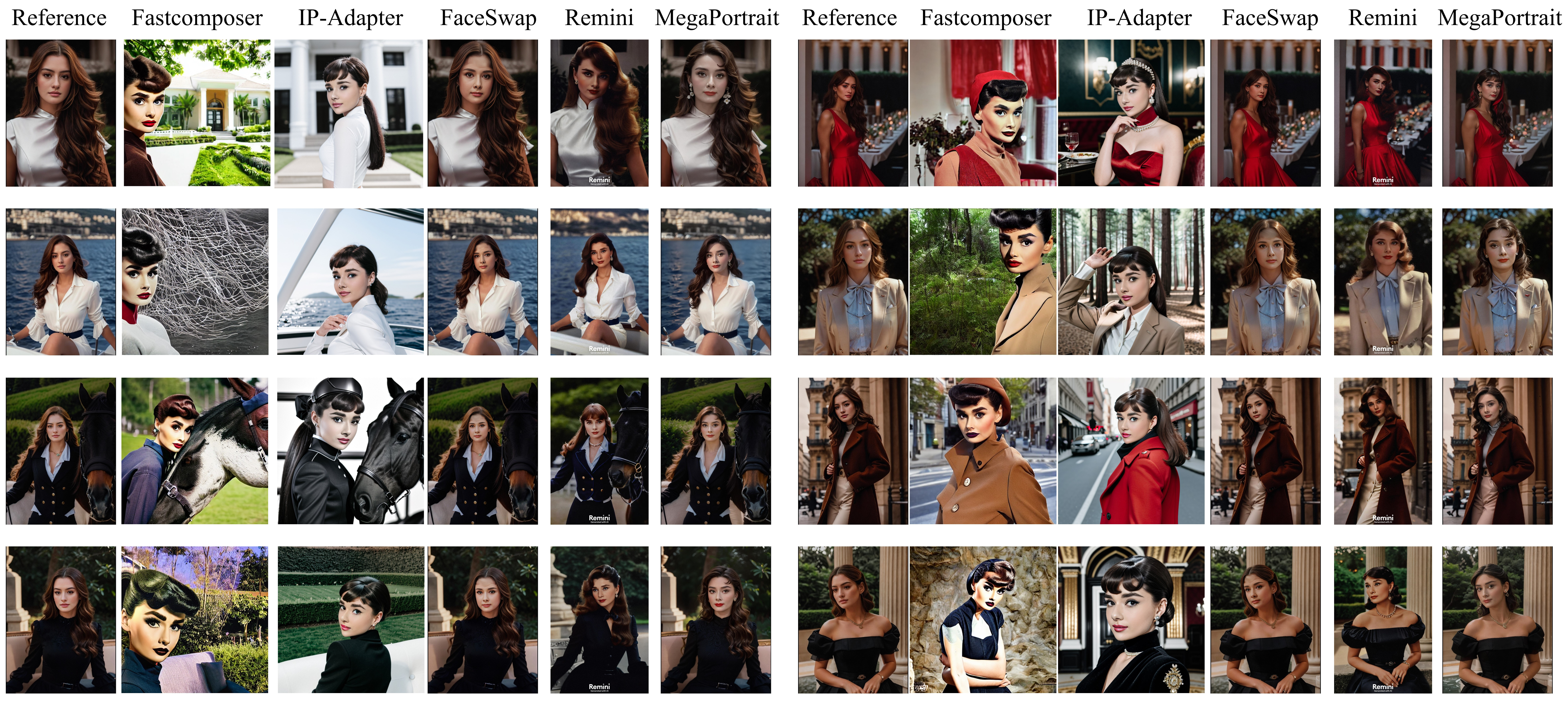}

   \caption{The visual comparison of MegaPortrait with the state-of-the-art portrait generation baseline methods, Fastcomposer~\cite{FastComposer} and IP-Adapter~\cite{IP-Adapter}, by using Audrey Hepburn as source ID. Our results can keep the style to the maximum while preserving the identity of the source image. Since Fastcomposer and IP-adapter only support ID-reference with prompting, we extract the prompts from reference image to test these methods. We also compare our results with the state-of-the-art AI photoshooting product Remini~\cite{remini}. The FaceSwap baseline adopts the swapper from insightface~\cite{insightface}.}
   \label{fig:qualitative}
\end{figure*}

For classifier free guidance, the conditional and unconditional diffusion scores are linearly combined as:
\begin{equation}
\tilde{\epsilon}_{\theta}(z_{\lambda}, c) = (1 + w){\epsilon}_{\theta}(z_{\lambda}, c) - w{\epsilon}_{\theta}(z_{\lambda}),
\end{equation}
where $\epsilon_{\theta}(z_{\lambda})$ is the unconditional score.

To affect the diffusion process, we introduce the same technique in classifier-free guidance by defining a new concept $S$ which is the textual description guiding the person looking at the camera. Practically, the concept $S$ can be customized and we find visually looking at the camera is the best practice to refine the gaze. We used the opposite concept $\Bar{S}$ for the negative score.

\begin{equation}
\tilde{\epsilon}_{\theta}(z_{\lambda}, S) = (1 + w){\epsilon}_{\theta}(z_{\lambda}, S) - w{\epsilon}_{\theta}(z_{\lambda},\Bar{S}).
\end{equation}
The final diffusion score is the combination of these two guidance. We introduce a local mask $M$ (the eyes mask extracted by dlib~\cite{dlib}) to restrict the editing area,
\begin{equation}
    \tilde{\epsilon}_{\theta}=\tilde{\epsilon}_{\theta}(z_{\lambda}, c)+M \odot \tilde{\epsilon}_{\theta}(z_{\lambda}, S).
\end{equation}
Practically, the extra diffusion score can be accumulated as a momentum with exponential averaging and decay the term along with the timestamps.

\section{Experiments}
\noindent \textbf{Implementation Details}. We rely on the utilization of the pre-trained Stable Diffusion V1.5 model. The personalization training phase is conducted over a total of 1500 steps, employing LoRa optimization with a learning rate of 1.5e-4. To optimize memory usage, we adopt the Adam8bit optimizer. The training process is executed on a single GeForce RTX A5000 GPU, with a batch size of 1. Throughout our experiments, we consistently utilize a training dataset comprising 10 to 13 images.

\noindent \textbf{Baselines and Metric}. 
In our comparative analysis, we employ both qualitative and quantitative methodologies to assess the performance of MegaPortrait against state-of-the-art AI human generators, namely Fastcomposer and IP-Adapter-FaceID. Additionally, we compare MegaPortrait with Remini, a widely-used AI photo-shooting application that ranks among the top three in downloads on both Google Play and the iOS App Store.

For quantitative evaluations, we meticulously select four identities, primarily sourced from the internet, with a focus on celebrities. Furthermore, we manually curate 25 style concepts, resulting in 100 id-style pairs for evaluation purposes. We gauge the facial appearance similarity between the source and generated portrait images by computing the cosine similarity between their facial embeddings, utilizing the arcface~\cite{arcface} technique. To evaluate the aesthetic quality of the generated portraits, we employ the Aesthetic Score Predictor V2 from Laion5B.

In addition to quantitative assessments, we conduct qualitative comparisons to provide a comprehensive understanding of the visual fidelity and artistic coherence achieved by MegaPortrait in comparison to its counterparts. Through these rigorous evaluations, we aim to provide insights into the strengths and limitations of MegaPortrait, establishing its position as a leading solution in the domain of personalized portrait generation within the realm of artificial intelligence.

\subsection{Visual Comparison}

The visual comparison is depicted in Fig.\ref{fig:qualitative} and Fig.\ref{fig:multiple}. Audrey Hepburn serves as the source identity for all methods under evaluation. Notably, as Fastcomposer~\cite{FastComposer} and IP-Adapter~\cite{IP-Adapter} lack support for reference style, we resort to extracting prompts from the reference image utilizing Blip~\cite{Blip}.

In Fig.~\ref{fig:qualitative}, it is discernible that our method yields results comparable to those achieved by Remini. Fastcomposer exhibits difficulties in generating coherent outcomes and fails to effectively capture and preserve identity information. Although IP-Adapter performs relatively better, the consistency of identity is not upheld to a satisfactory degree upon human visual assessment. In contrast, MegaPortrait excels in preserving identity to the utmost extent while concurrently retaining the reference style. Notably, the synthesized results produced by MegaPortrait demonstrate the highest degree of consistency with human-perceived identity information.

\subsection{Quantitative Results}

\begin{table}[htbp]
  \centering
  \caption{The quantitative results with the baseline methods. For ID score we use the arcface~\cite{arcface} identity extractor and measure by cosine similarity. For Aesthetic score we adopt the Aesthetic Score Predictor V2~\cite{Laion5B} from Laion5B.}
    \begin{tabular}{lll}
    \toprule
    \textbf{Model} & \textbf{ID Score} & \textbf{Aesthetic Score} \\
    \midrule
    Fastcomposer~\cite{FastComposer} & 0.3802 & 5.7382 \\
    IP-Adapter-FaceId~\cite{IP-Adapter} & 0.6692 & 6.1062 \\
    Remini~\cite{remini} & 0.6157 & 6.2431 \\
    MegaPortrait & 0.6630 & \textbf{6.3789} \\
    \bottomrule
    \end{tabular}%
  \label{tab:addlabel}%
\end{table}%

The quantitative results are presented in Table~\ref{tab:addlabel}, encompassing two key metrics: ID Score and Aesthetic Score. The ID Score is evaluated based on the cosine similarity computed using the arcface extractor~\cite{arcface}, while the Aesthetic Score is determined utilizing the aesthetic predictor from Laion5B~\cite{Laion5B}.

Our method consistently outperforms Remini across both metrics, achieving superior results in terms of ID Score and Aesthetic Score. Furthermore, our method demonstrates comparable performance to IP-Adapter, highlighting its efficacy in generating high-quality portrait images with respect to both identity preservation and aesthetic appeal.

It is worth noting that while the ID Score provides valuable quantitative insights, it may have limitations in capturing nuances of human-perceived evaluation. Nonetheless, our method showcases strong alignment as evidenced by the multiple people visual results depicted in Fig.~\ref{fig:multiple}. These findings underscore the effectiveness of our approach in achieving a harmonious balance between quantitative metrics and qualitative human assessment.

\begin{figure}[th]
  \centering
   \includegraphics[width=0.7\linewidth]{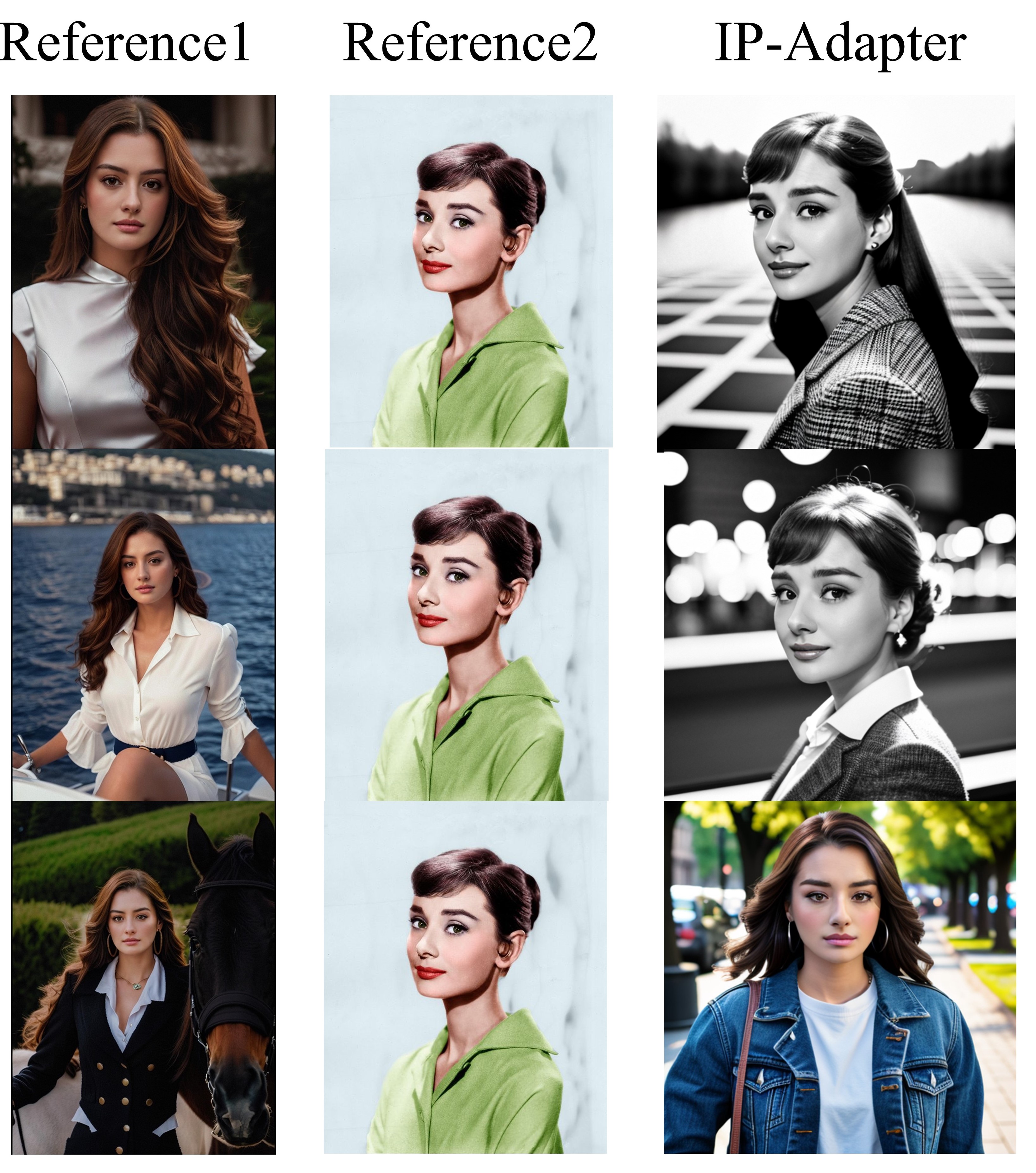}

   \caption{The results of using IP-Adapter-FaceID with style and identity reference. IP-Adapter-FaceID can only handle with identity input and discards the style information. Meanwhile, the IP-Adapter~\cite{IP-Adapter} original version only supports style reference and lacks in ability to capture identity information, as shown in its paper. }
   \label{fig:IPA_multi}
\end{figure}

\begin{figure}[th]
  \centering
   \includegraphics[width=1\linewidth]{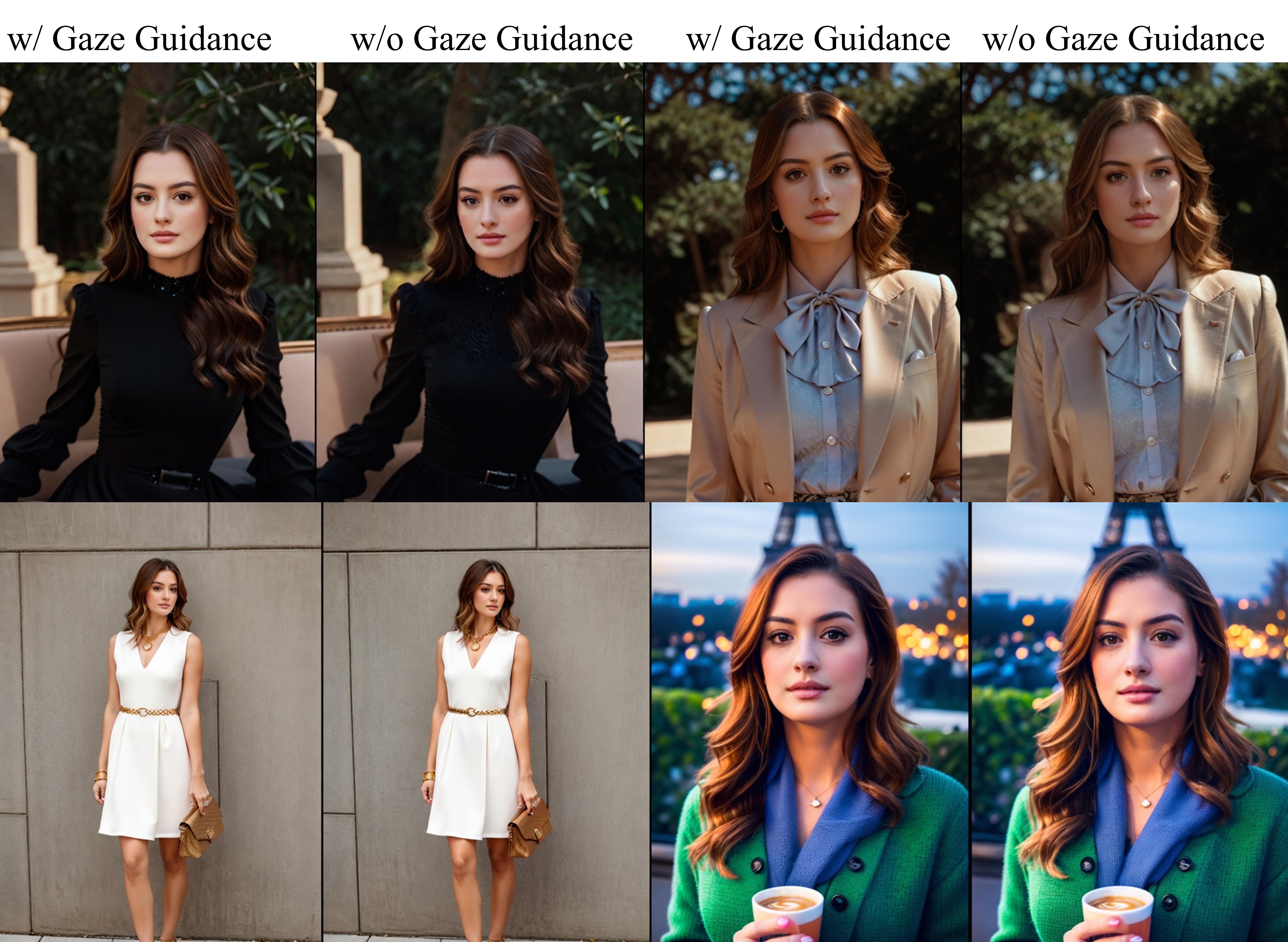}

   \caption{The visual comparison of MegaPortrait using Gaze Guidance or not. The portrait image with Gaze Guidance has 
 a better gaze direction towards the viewer, which is a more vivid setting aesthetically. The guidance prompt used here is ``looking at the camera".}
   \label{fig:gaze}
\end{figure}

\begin{figure}[th]
  \centering
   \includegraphics[width=1\linewidth]{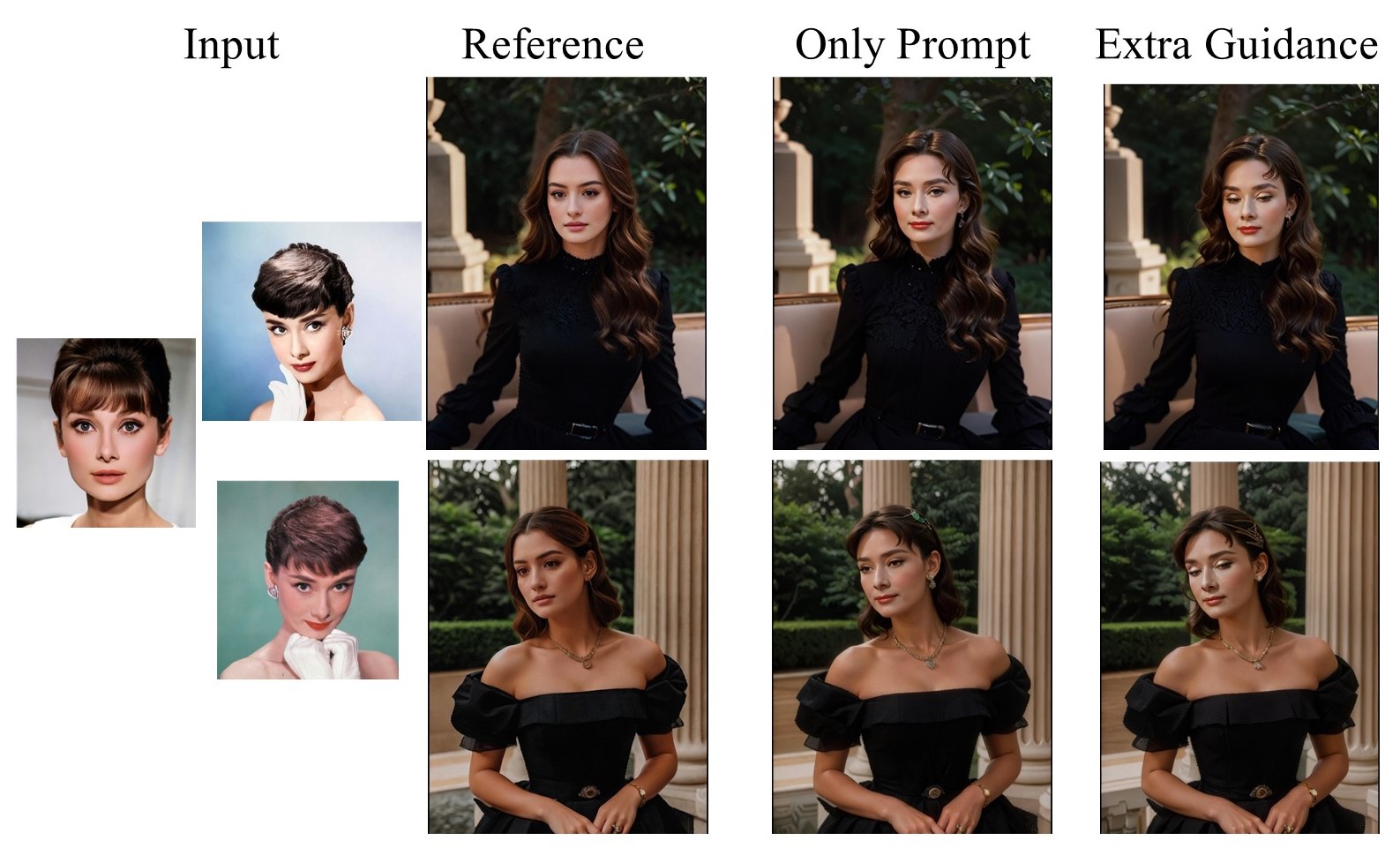}

   \caption{We demonstrate another setting with our extra guidance by closing the model's eyes. The extra guidance actually works as a local editor, not just an ad-hoc gaze corrector. By only adding ``Closed eyes" on prompts, the results are not generated correctly. }
   \label{fig:closed_eyes}
\end{figure}

\begin{figure}[th]
  \centering
   \includegraphics[width=1\linewidth]{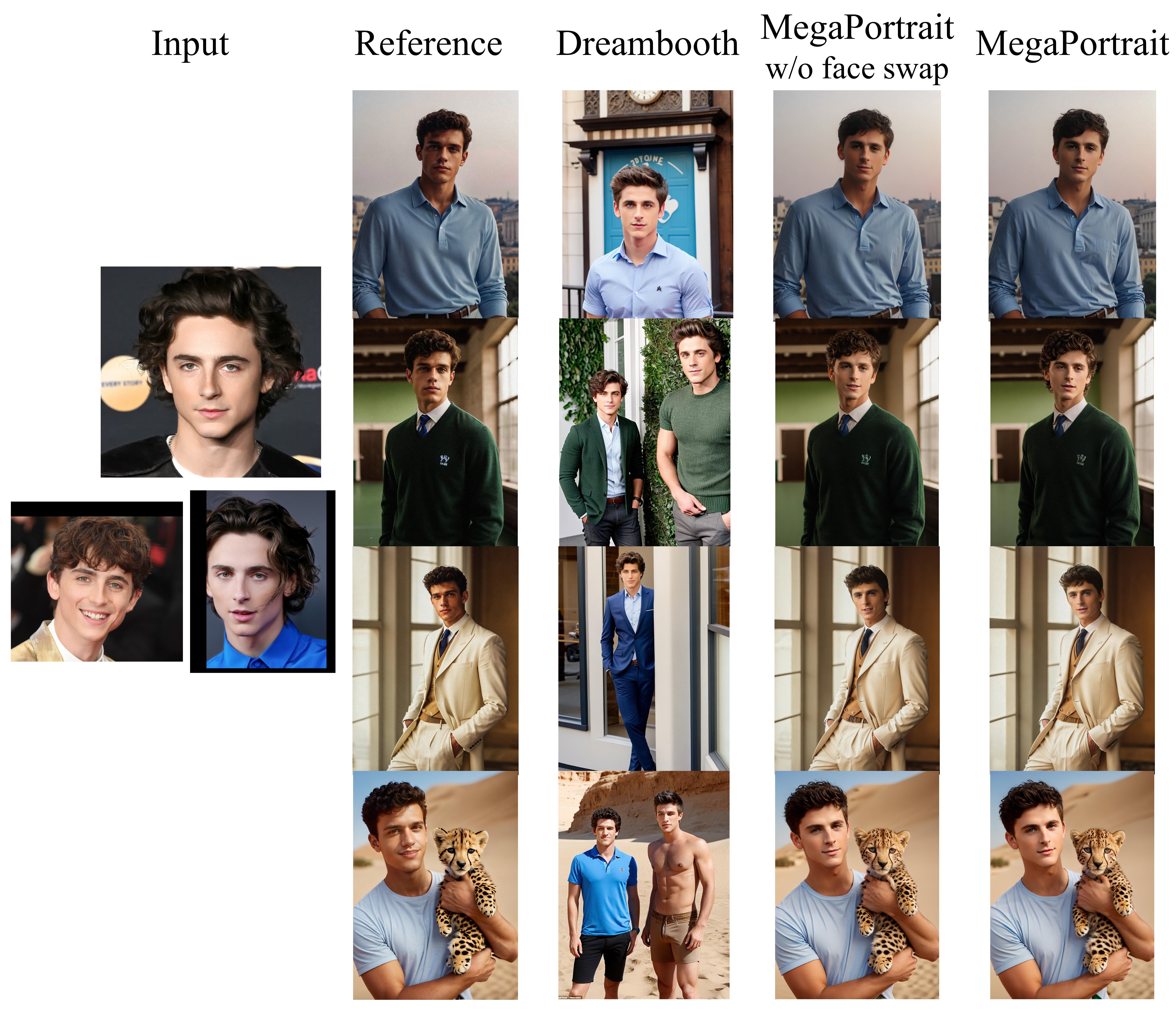}

   \caption{Ablation study with dreambooth~\cite{Dreambooth} and a MegaPortrait variation without face swap.}
   \label{fig:dreambooth}
\end{figure}

\subsection{Ablation Study}
\noindent \textbf{Adding style reference to IP-Adapter}.\quad
In Fig.~\ref{fig:IPA_multi}, we provide a demonstration highlighting the limitations of directly feeding multiple image prompts to IP-Adapter, resulting in the blending of styles rather than distinct representations.

\noindent \textbf{MegaPortrait v.s. Dreambooth}.\quad
Moving on to Fig.~\ref{fig:dreambooth}, we conduct a comparative analysis between MegaPortrait and variations that exclude face swap in the intermediate results. Here, face swap serves as an optional identity enhancer. Additionally, we compare MegaPortrait with Dreambooth~\cite{Dreambooth}. While Dreambooth excels in identity preservation, it falls short in retaining style references effectively.

\noindent \textbf{Gaze Improvement.}\quad
Furthermore, in Fig.~\ref{fig:gaze}, we showcase the effectiveness of incorporating additional gaze guidance, resulting in more vivid and expressive gazes within the generated portraits. Moreover, we demonstrate the versatility of our approach as a general post-processing editor in Fig.~\ref{fig:closed_eyes}, where we effectively modify the gaze by simulating closed eyes.

These visual demonstrations underscore the versatility and effectiveness of MegaPortrait across various scenarios, showcasing its capabilities in identity preservation, style fidelity, and post-processing enhancements.

\section{Conclusion \& Limitations}
In summary, we present MegaPortrait, an innovative system for personalized portrait generation that effectively balances the preservation of identity with adaptation to a variety of artistic styles. By revisiting diffusion control conditions and incorporating advanced techniques such as geometric and shadow decoupling and gaze correction guidance, MegaPortrait achieves state-of-the-art results in terms of visual fidelity and aesthetic quality.

Through an elaborate pipeline consisting of Identity Net, Shading Net, and Harmonization Net, MegaPortrait seamlessly integrates each character's unique features with specified styles and poses extracted from reference images. The use of a diffusion model enables MegaPortrait to generate high-fidelity and visually compelling personalized portrait images with remarkable consistency in identity preservation and style fidelity.

Quantitative evaluation and qualitative comparisons with leading AI portrait methods demonstrate MegaPortrait's superiority in terms of identity preservation, style fidelity, and overall aesthetic appeal. Additionally, ablation studies further highlight MegaPortrait's effectiveness and versatility in a variety of scenarios, including gaze correction and post-processing enhancement. We also compared it with the leading closed-source product Remini and demonstrated comparable or better performance.

For limitations, MegaPortrait cannot preserve the original hair color of the source ID. It also may fail in generating correct eye colors as given in Fig.~\ref{fig:multiple} column 1. 

Future research can work on the following directions: 1) Aesthetic level mining. Since directly embedding the aesthetic level into the framework is ad-hoc, we can propose a method to mine the aesthetic displacement with user-edited portraits which can be used to adapt to different culture zones and preferences. 2) We can train a non-spatial controlnet which can perform the exactly same task without and training. It might be a better design choice than encoder-based methods. 3) The facial expressions are coupled with identity, better strategies can be applied to generate various expressions.


{\small
\bibliographystyle{ieee_fullname}
\bibliography{egbib}
}

\end{document}